\documentclass[lettersize,journal]{IEEEtran}
\usepackage{amsmath,amsfonts}
\usepackage{makecell}
\usepackage{array}
\usepackage[caption=false,font=normalsize,labelfont=sf,textfont=sf]{subfig}
\usepackage{textcomp}
\usepackage{stfloats}
\usepackage{url}
\usepackage{verbatim}
\usepackage{graphicx}
\usepackage{cite}
\usepackage{bigstrut}
\usepackage{multirow}
\usepackage{bm}
\usepackage{algpseudocode} 
\usepackage{booktabs}
\usepackage[ruled]{algorithm}
\usepackage{float}
\usepackage{graphicx}
\usepackage{epstopdf}
\usepackage{setspace}
\usepackage{amssymb}
\begin{document}

\title{Fuzzy Knowledge Distillation from \\High-Order TSK to Low-Order TSK}

\author{Xiongtao Zhang, Zezong Yin, Yunliang Jiang, Yizhang Jiang, Danfeng Sun and Yong Liu
\thanks{This work was supported in part by National Natural Science Foundation of China (U22A20102, 62171203), in part by the “Pioneer” and “Leading Goose” R \& D Program of Zhejiang Province (2023C01150), and in part by the open project fund of Key Laboratory of Image Processing and Intelligent Control (Huazhong University of science and technology), Ministry of Education (\emph{Corresponding author: Yunliang Jiang}).}
\thanks{Xiongtao Zhang and Zezong Yin are with the Zhejiang Province Key Laboratory of Smart Management\&Application of Modern Agricultural Resources, Huzhou University, Huzhou 313000, China, and also with the School of Information Engineering, Huzhou University, Huzhou 313000, China.}
\thanks{Yunliang Jiang is with the School of Computer Science and Technology, Zhejiang Normal University, Jinhua 321004, China, and also with the School of Information Engineering, Huzhou University, Huzhou 313000, China. Yizhang Jiang is with the School of Artificial Intelligence and Computer Science, Jiangnan University, Wuxi 214122, China, and also with the Key Laboratory of Image Processing and Intelligent Control (Huazhong University of Science and Technology), Ministry of Education, Wuhan 430074, China. Danfeng Sun is with the School of Computer Science, Hangzhou Dianzi University, Hangzhou 310018, China. Yong Liu is with the College of Control Science and Enginneering, Zhejiang University, Hangzhou 310007, China.}}

\markboth{Journal of \LaTeX\ Class Files,~Vol.~14, No.~8, August~2021}%
{Shell \MakeLowercase{\textit{et al.}}: A Sample Article Using IEEEtran.cls for IEEE Journals}

\maketitle

\begin{abstract}
High-order Takagi-Sugeno-Kang (TSK) fuzzy classifiers possess powerful classification performance yet have fewer fuzzy rules, but always be impaired by its exponential growth training time and poorer interpretability owing to High-order polynomial used in consequent part of fuzzy rule, while Low-order TSK fuzzy classifiers run quickly with high interpretability, however they usually require more fuzzy rules and perform relatively not very well. Address this issue, a novel TSK fuzzy classifier embeded with knowledge distillation in deep learning called HTSK-LLM-DKD is proposed in this study. HTSK-LLM-DKD achieves the following distinctive characteristics: 1) It takes High-order TSK classifier as teacher model and Low-order TSK fuzzy classifier as student model, and leverages the proposed LLM-DKD (Least Learning Machine based Decoupling Knowledge Distillation) to distill the fuzzy dark knowledge from High-order TSK fuzzy classifier to Low-order TSK fuzzy classifier, which resulting in Low-order TSK fuzzy classifier endowed with enhanced performance surpassing or at least comparable to High-order TSK classifier, as well as high interpretability; specifically 2) The Negative Euclidean distance between the output of teacher model and each class is employed to obtain the teacher logits, and then it compute teacher/student soft labels by the softmax function with distillating temperature parameter; 3) By reformulating the Kullback-Leibler divergence, it decouples fuzzy dark knowledge into target class knowledge and non-target class knowledge, and transfers them to student model. The advantages of HTSK-LLM-DKD are verified on the benchmarking UCI datasets and a real dataset \emph{Cleveland heart disease}, in terms of classification performance and model interpretability.
\end{abstract}

\begin{IEEEkeywords}
Deep learning, Fuzzy dark knowledge, High-order Takagi-Sugeno-Kang (TSK) fuzzy classifier, Knowledge distillation, Least Learning Machine.
\end{IEEEkeywords}

\section{Introduction}
\IEEEPARstart{T}{Akagi-Sugeno-Kang} (TSK) fuzzy classifier is one of the most famous fuzzy classifiers, which consists of antecedent part and consequent part with fuzzy rule. The antecedent part divides the input space into several fuzzy areas, the consequent part describes the logic of classifier in those areas. TSK fuzzy classifier has been deeply combined with deep learning \cite{ref1,ref2,ref3}, and successfully applied to many application fields, including epileptic seizure detection \cite{ref4,ref5,ref6}, subway fare pricing \cite{ref7}, and vehicles path planning \cite{ref8}.
\par
Among a wide variety of TSK fuzzy classifiers, due to their high interpretability and fast training speed, Zero-order \cite{ref9} and First-order TSK fuzzy classifiers \cite{ref10} have attracted the most 
widespread attentions. However, they are prone to resulting in \emph{rule explosion} \cite{ref40}, which means the number of fuzzy rules grows rapidly when desire the enhanced classification performance. Address this issue, many methods have been developed, which can be classified into three categories:

\begin{enumerate}
	\item{$\emph{Hierarchical TSK fuzzy classifier}$, organizes each layer in a stacked way based on stacking generation principle \cite{ref11} to obtain improved classification performance. Xiongtao Zhang \emph{et al.} \cite{ref3} proposed a novel hierarchical-structure of TSK fuzzy subclassifiers called EP-TSK-FK, which quickly built subclassifiers in terms of parallel learning to obtain augmented validation data, and thus got prediction by fuzzy clustering and KNN. Wang \emph{et al.} \cite{ref40} invented a hierarchical TSK fuzzy classifier with shared linguistic fuzzy rules by opening the manifold structure of the original input space, in which the input of each layer includes the output of previous layer besides the original data.}
	\item{$\emph{Deep learning based TSK fuzzy classifier}$, integrates the deep learning strategies with TSK fuzzy classifier. Wu \emph{et al.} \cite{ref12} extended Dropout to fuzzy rule, where the model dropped some rules at random during training, and used all rules during testing. Cui \emph{et al.} \cite{ref13} avoided the excessive influence of strong rules by adjusting the membership degree of fuzzy rules with normalization.}
	\item{$\emph{High-order TSK fuzzy classifier}$, uses High-order polynomial in the consequent part to escape the influence of \emph{rule explosion} and achieve better classification performance with fewer rules. Ren \emph{et al.} \cite{ref41}  built a High-order type-2 TSK system, in which the membership functions in antecedent part are type-2 fuzzy sets and the consequent part are High-order polynomial function. However, High-order TSK fuzzy classifier still has serious shortcoming: the interpretability suffers substantially damage since the parameters of High-order polynomial in consequent part are overly complex.}
\end{enumerate}

In deep learning, large model usually has strong performance owing to using a huge amount of computation to extract structure from data, whose complex parameters makes it difficult to be applied in various real-world fields. In contrast, small model is deployed easily but it performs not very well in practice. Inspired by this, Hinton \emph{et al.} \cite{ref14} proposed model compression technique knowledge distillation in recent years, which compresses a student model (usually small or simple model) from a teacher model (usually large or complex model), guides the student model by transferring the dark knowledge via minimizing the Kullback-Leibler divergence (KL divergence) between prediction logits of teacher model and student model, thus improving the performance of student model. As research continues, the direction of knowledge distillation is growing increasingly broad. There have been various knowledge forms and distillation methods in knowledge distillation \cite{ref15,ref16,ref17,ref18,ref19,ref20,ref21}. The technique of knowledge distillation are also widely used, such as object recognition \cite{ref22,ref23,ref24}, defect detection \cite{ref25,ref26} and other fields \cite{ref27,ref28,ref29}, etc. Recent studies have found that knowledge distillation can greatly improve the performance of TSK fuzzy classifier. Gu \emph{et al.} \cite{ref30} transfered knowledge from CNN to TSK fuzzy classifier and then explained how the TSK fuzzy classifier made decisions. Erdem \emph{et al.} \cite{ref31} used CNN to distill interval type-2 fuzzy classifier, which improved the classification performance on large datasets. Our previous study \cite{ref1} born-again TSK fuzzy classifier with CNN, which took dark knowledge from CNN as parameters of antecedent part and consequent part of TSK fuzzy classifier respectively, then expressed dark knowledge in an interpretable manner.
\par
As we well known, High-order TSK fuzzy classifier has been exhibiting its outstanding classification performance because of powerful fitting ability, while Low-order TSK fuzzy classifiers have demonstrated strength in concise interpretability due to its interpreatble consequent part. In this paper, we extend our previous study \cite{ref1} from only binary-class classification task to multi-class classification task by born-again TSK fuzzy classifier using fuzzy knowledge distillation, and propose a novel TSK fuzzy classifier called HTSK-LLM-DKD, which decouples and transfers fuzzy dark knowledge from High-order TSK fuzzy classifier (teacher model) to Low-order TSK fuzzy classifier (student model) while obtaining better classification performance as well as high interpretability. Our study advances in the combination of knowledge distillation and TSK fuzzy classifier to a deeper level. The contributions of this study can be summarized as follows: 
\par
\begin{enumerate}
	\item{HTSK-LLM-DKD proposes a novel least learning machine based decoupling knowledge distillation, denoted as LLM-DKD. By comparison with the gradient descent approach, HTSK-LLM-DKD can quickly solve consequent part of teacher model by Least Learning Machine (LLM) \cite{ref2}. Furthermore, LLM-DKD uses the Negative Euclidean distance between output and each class to obtain logits. According to \cite{ref32}, logits in LLM-DKD can represent more comprehensive class information and transfer fuzzy dark knowledge better with higher semantic level.}
	\item{HTSK-LLM-DKD employs the \emph{softmax} function with distillating temperature parameter to obtain soft labels of teacher model and student model. With reformulating the Kullback-Leibler divergence (KL divergence), HTSK-LLM-DKD decouples fuzzy dark knowledge into target class knowledge and non-target class knowledge, providing more flexible and efficient logits distillation perspective, so as to further improve the performance of TSK fuzzy classifier.}
	\item{Experimental results on benchmarking datasets and a real-world dataset \emph{Cleveland Heart Disease} demonstrate the effectiveness of the proposed HTSK-LLM-DKD. In terms of classification performance, HTSK-LLM-DKD achieves the best performance on most datasets and perform better than High-order TSK fuzzy classifier with powerful generalization ability; in terms of interpretability, HTSK-LLM-DKD is inherently comparable with high interpretable Low-order TSK fuzzy classifiers.}
\end{enumerate}

Table S1 summarizes the notations used in our study in the supplementary file.

\section{Related Work}
\subsection{TSK fuzzy classifier}
TSK fuzzy classifier can be described by fuzzy IF-THEN rules, which indicate the input-output relationship of the classifier. For TSK fuzzy classifiers, the fuzzy rules can be expressed as follows:
\begin{align*}
	\label{eq1a}
	&{\rm IF} \ x_{1} \ {\rm is} \ A_{1}^{k} \wedge x_{2} \ {\rm is} \ A_{2}^{k} \ \wedge \ldots \wedge \ x_{m} \ {\rm is} \ A_{m}^{k}\\
	&{\rm THEN} \
	y^{k}=f^{k}(\mathbf{x})\tag{1a} 
\end{align*}where $k=1,2,\ldots,K$. $K$ is the total number of fuzzy rules, the input is denoted as $\mathbf{x}=(x_{1},x_{2},\ldots,x_{i},\ldots,x_{m})^{T}$, $A_{i}^{k}$ is the fuzzy set of the $k$-th rule of $x_{i}$, $y^{k}$ is the output of the $k$-th rule, $\wedge$ is the fuzzy conjunction operation. If $f^{k}(\mathbf{x})=p_{0}^{k}$, we term the TSK fuzzy classifier as the
Zero-order TSK fuzzy classifier \cite{ref9}. If $f^{k}(\mathbf{x})=p_{0}^{k}+x_{1}^{k}p_{1}^{k}+x_{2}^{k}p_{2}^{k}+\ldots+x_{m}^{k}p_{m}^{k}$, the TSK fuzzy classifier is termed as the First-order TSK fuzzy classifier \cite{ref10}. 	If $f^{k}(\mathbf{x})$ is described as:
\begin{equation}
	\label{eq1b}
	f^{k}(\mathbf{x})=\sum_{\substack{j_{1}+j_{2}+\ldots+j_{m}\le n \\j_{1},j_{2},\ldots,j_{m}\ge 0}} a_{j_{1},j_{2},\ldots,j_{m}}^{k}x_{1}^{j_{1}}x_{2}^{j_{2}}\ldots x_{m}^{j_{m}}\tag{1b}
\end{equation}It is termed as High-order TSK fuzzy classifier \cite{ref11}. Where $n (n \geq 2)$ is the order of the highest polynomial of TSK fuzzy classifier, $j_{i}$ is the order of $x_{i}$, $a_{j_{1},j_{2},\ldots,j_{m}}^{k}$ represents the coefficient of the highest polynomial with $m$ independent variables in the linear combination constituting the $k$-th rule.
\par
It is clearly that the consequent parameters of Zero-order and First-order TSK fuzzy classifier are relatively simple, they are the coefficients of the fuzzy membership degree and input variables, which ensure the high interpretability but need more fuzzy rules to achieve satisfactory performance. In contrast, High-order TSK fuzzy classifier can preform better with fewer fuzzy rules, but the consequent part of fuzzy rule are extremely complex and sharply reduce the interpretability. As a result, how to combine the strong performance of High-order TSK fuzzy classifier with the high interpretability of Low-order TSK fuzzy classifier becomes a very valuable research direction.
\subsection{Knowledge Distillation}
\begin{figure}[!t]
	\centering
	\includegraphics[width=3.7in]{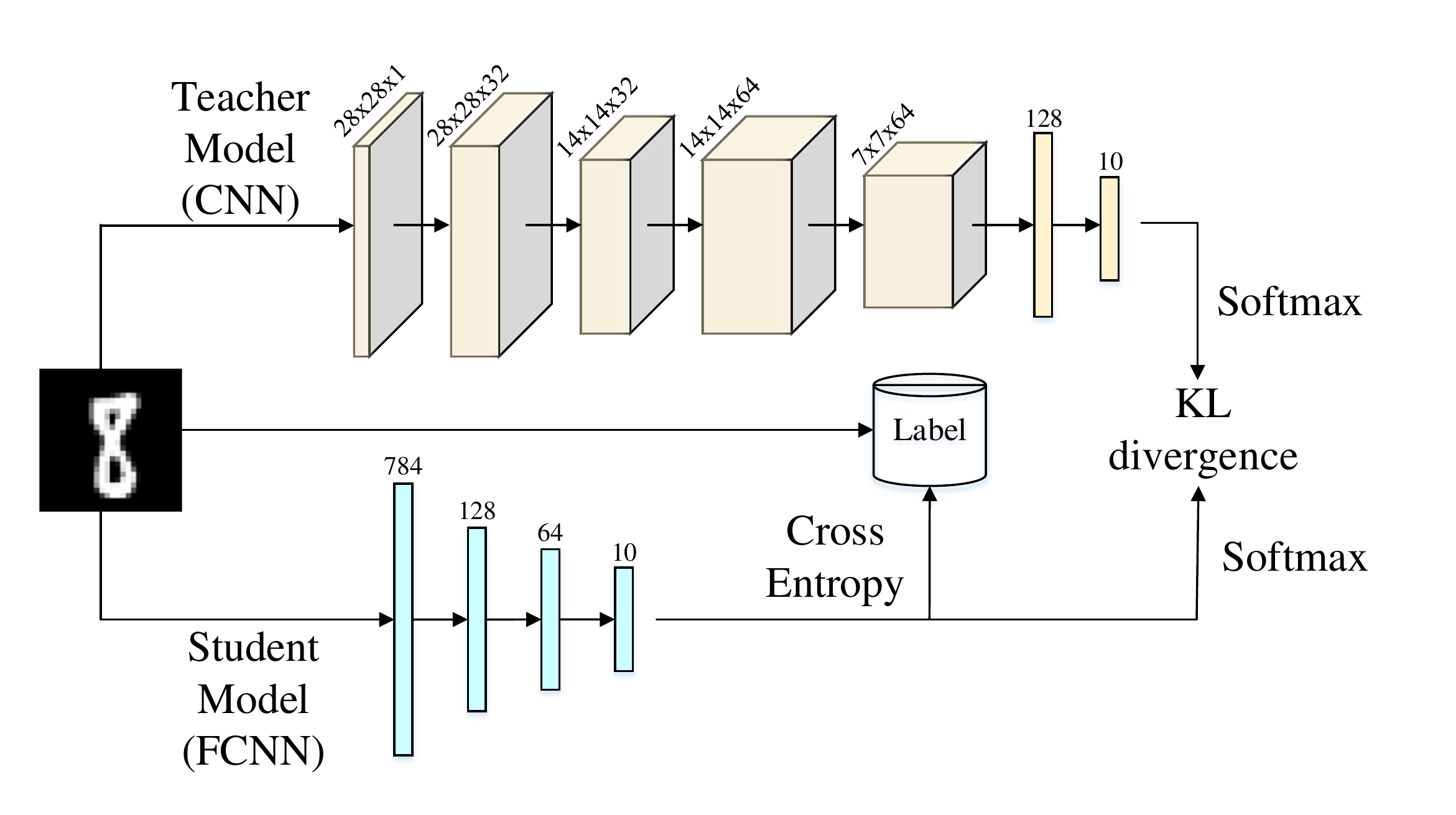}
	\caption{Architecture of knowledge distillation. Taking Convolutional Neural Network (CNN) as teacher model and Fully Connected Neural Network (FCNN) as student model for example.}
	\label{fig_1}
\end{figure}

Knowledge distillation \cite{ref14} transfers the dark knowledge from complex/large model, namely the teacher model, to simple/small model, namely the student model, and hence improve the performance of student model. Specifically, knowledge distillation introduces the hyper-parameter temperature in \emph{softmax} function to obtain soft labels at first, then calculates the KL divergence of soft labels and the cross-entropy of student model output and the ground-truth label, finally transfers the dark knowledge via soft labels from teacher model to student model, improves the performance of student model, as shown in Fig.~\ref{fig_1}.
\par
There are many strategies to construct the loss function in knowledge distillation, such as the KL divergence \cite{ref14}, the mean squared error \cite{ref37} and the Jensen–Shannon divergence \cite{ref38}, etc. Traditional knowledge distillation transfers dark knowledge in a highly coupled way, which limits the flexibility for knowledge transfer. Zhao \emph{et al.} \cite{ref32} pointed out that dark knowledge can be decoupled into target class knowledge and non-target class knowledge and to be transfered to student model in a more flexible way by reconstructing the KL divergence. Furlanello \emph{et al.} \cite{ref33} demonstrated that the decoupled dark knowledge of teacher model can guide student model to have stronger generalization ability than that of teacher model. In this paper, we attempt to distill fuzzy dark knowledge from High-order TSK fuzzy classifier, and propose a novel born-again TSK fuzzy classifier endowed with the powerful classification performance as well as high interpretability. 

\section{Htsk-llm-dkd}

\begin{figure*}[!t]
	\centering
	\includegraphics[width=7in]{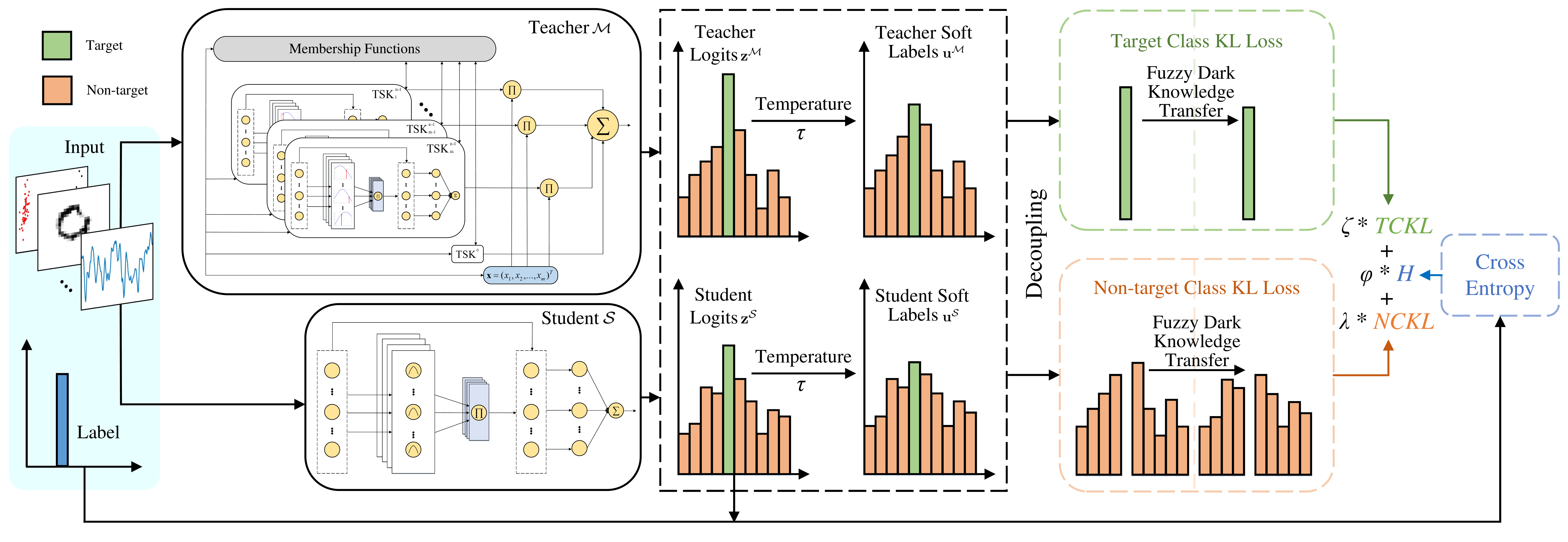}
	\caption{Architecture of HTSK-LLM-DKD. We distill fuzzy dark knowledge from High-order TSK fuzzy classifier to Low-order TSK fuzzy classifier, specifically decouple KL divergence into Target Class KL Loss (\emph{TCKL}) and Non-Target Class KL Loss (\emph{NCKL}), optimize the proposed HTSK-LLM-DKD with \emph{TCKL}, \emph{NCKL} and Cross-Entropy (\emph{H}) as loss function.}
	\label{fig_2}
\end{figure*}

In this section, we propose a novel TSK fuzzy classifier called HTSK-LLM-DKD, which distilling knowledge from High-order TSK fuzzy classifier (teacher model) to Low-order TSK fuzzy classifier (student model), utilizing the proposed LLM-DKD. Specifically, LLM-DKD firstly trains teacher model quickly and takes the Negative Euclidean distance between the output value and each class to obtain teacher logits. Then, it uses the \emph{softmax} function with temperature parameter $\tau$ to obtain soft labels of teacher model and student model, respectively. Finally, LLM-DKD decouples fuzzy dark knowledge into target class knowledge and non-target class knowledge, and hence transfers them to student model efficiently, so as to obtain better classification performance as well as high interpretability. The overall architecture of HTSK-LLM-DKD is shown in Fig.~\ref{fig_2}.
\subsection{Specific Architecture of Teacher Model and Student Model}
\subsubsection{Constructing Process of Teacher Model}
Based on the definition of High-order polynomial, we prove that High-order polynomial can be composed of several Low-order polynomials, the mathematical proof process is present in the supplementary file. In this paper, Third-order TSK fuzzy classifier is employed as teacher model, which is built by stacking multiple zero-order TSK fuzzy classifiers \cite{ref34}, \emph{i.e.}, the consequent part of teacher model can be expressed as the superposition of the consequent parts of several Low-order TSK fuzzy classifiers:
\begin{align*}
	\label{eq1c}
	y= \ &y_{0}+x_{1}\left(y_{0}^{(1,0)}+x_{1} y_{1}^{(1,1)}+x_{2} y_{1}^{(1,2)}+\ldots+x_{m} y_{1}^{(1, m)}\right)\\
	&+ x_{2}\left(y_{0}^{(2,0)}+x_{1} y_{1}^{(2,1)}+x_{2} y_{1}^{(2,2)}+\ldots+x_{m} y_{1}^{(2, m)}\right)\\
	&+\ldots+\\
	&x_{m}\left(y_{0}^{(m,0)}+x_{1} y_{1}^{(m, 1)}+x_{2} y_{1}^{(m, 2)}+\ldots+x_{m} y_{1}^{(m, m)}\right)\tag{1c}
\end{align*}where $y_{n}^{(i,j)}$ represents the output of the $j$-th $n$-order TSK fuzzy classifier, which is used to construct the $i$-th ($n+1$)-order TSK fuzzy classifier. Specifically, $y_{0}$ indicates the output of Zero-order TSK fuzzy classifier.

\subsubsection{Calculation Process of Teacher Model and Student Model}
(\ref{eq1c}) is usually calculated by weighted summation:
\begin{equation}
	\label{eq2a}
	y=\sum_{k=1}^{K} \frac{\mu^{k}(\mathbf{x})}{\sum_{k^{\prime}=1}^{K} \mu^{k^{\prime}}(\mathbf{x})} y^{k}=\sum_{k=1}^{K} \tilde{\mu}^{k}(\mathbf{x}) y^{k}\tag{2a}
\end{equation}where $\mu^{k}(\mathbf{x})$ and $\tilde{\mu}^{k}(\mathbf{x})$ are the membership degree and the normalized membership degree of the input $\mathbf{x}$ in the $k$-th fuzzy rule, respectively. Gaussian membership function is widely used to calculate fuzzy membership degree:
\begin{align}
	\label{eq2b}
	\mu^{k}(\mathbf{x})=\prod_{i=1}^{m} {\mu^{k}(x_{i})} \quad \mu^{k}(x_{i})=\exp \left(\frac{-\left(x_{i}-v_{i}^{k}\right)^{2}}{2 \delta_{i}^{k}}\right)\tag{2b}
\end{align}where ${\mu^{k}(x_{i})}$ is the membership degree of $x_{i}$. $v_{i}^{k}$ is the center of each fuzzy rule, which is randomly selected from $\{0, 0.25, 0.5, 0.75, 1\}$, and may have linguistic explanation: $\{\emph{very low}, \emph{low}, \emph{medium}, \emph{high}, \emph{very high}\}$, thus ensuring the interpretable antecedent part of the proposed HTSK-LLM-DKD. $\delta_{i}^{k}$ is the kernel width, which is set to be a positive value.

Several mathematical transformations from the fuzzy rule to linear function exist, as follows.
\begin{align}
	\label{eq3a}
	\mathbf{x}_{e} &= \left(1, \mathbf{x}^{T}\right)^{T} & \mathbf{x}_{f} &= \left(1, x_{i}\mathbf{x}_{e}^{T}\right)^{T} & \mathbf{x}_{v} &= \left(1, x_{i}\mathbf{x}_{f}^{T}\right)^{T}\tag{3a}
\end{align}
\begin{align}
	\label{eq3b}
	\tilde{\mathbf{x}}_{g}^{k} &= \tilde{\mu}^{k}(\mathbf{x})\mathbf{x}_{v}& \mathbf{x}_{g}&=\left[\left(\tilde{\mathbf{x}}_{g}^{1}\right)^{T},\left(\tilde{\mathbf{x}}_{g}^{2}\right)^{T},\ldots,\left(\tilde{\mathbf{x}}_{g}^{K}\right)^{T}\right]^{T}\tag{3b}
\end{align}
\begin{align}
	\label{eq3c}
	\tilde{\mathbf{x}}_{h}^{k} &= \tilde{\mu}^{k}(\mathbf{x})\mathbf{x}_{e}&	\mathbf{x}_{h}&=\left[\left(\tilde{\mathbf{x}}_{h}^{1}\right)^{T},\left(\tilde{\mathbf{x}}_{h}^{2}\right)^{T},\ldots,\left(\tilde{\mathbf{x}}_{h}^{K}\right)^{T}\right]^{T}\tag{3c}
\end{align}$\mathcal{M}$ and $\mathcal{S}$ are denoted as teacher model and
student model, respectively. Our teacher model is Third-order TSK fuzzy classifier, which performs well but usually spends much time to train, in this paper we use Least Learning Machine (LLM) \cite{ref2} to quickly solve the consequent part of teacher model:
\begin{align}
	\label{eq4}
	y^{\mathcal{M}}&=\mathbf{q}_{g}^{T}\mathbf{x}_{g} &	\mathbf{q}_{g} &= \left((1 / L) \mathbf{I}+\mathbf{X}_{g}^{T} \mathbf{X}_{g}\right)^{-1} \mathbf{X}_{g}^{T} \overline{\mathbf{Y}}\tag{4}
\end{align}where $y^{\mathcal{M}}$ is the output of teacher model, $\mathbf{q}_{g}$ is the consequent parameters of teacher model, $L$ is the regularization parameter, $\mathbf{X}_{g}=\left[\mathbf{x}_{g}^{1},\mathbf{x}_{g}^{2},\ldots,\mathbf{x}_{g}^{N}\right]^{T}$ , $\mathbf{I}$ is the identity matrix of $N\times N$, $N$ is the number of samples,  $\overline{\mathbf{Y}}=\left[\overline{Y}_{1},\overline{Y}_{2},\ldots,\overline{Y}_{N}\right]^{T}$ is the ground-truth label.

First-order TSK fuzzy classifier is utilized as student model in this paper, which runs fastly with high interpretability but indeed performs not very well compared with High-order TSK fuzzy classifier. In HTSK-LLM-DKD, the consequent parameters of student model can be updated by the gradient descent approach that is derived by minimizing the cross-entropy error criterion:
\begin{align}
	\label{eq5}
	\mathbf{z}^{\mathcal{S}}&=\mathbf{Q}_{h}^{T}\mathbf{x}_{h} & H &= -\sum_{i = 1}^{N}\sum_{t=1}^{C} \overline{Y}_{i,t} \log \left(z_{i,t}^{\mathcal{S}}\right)\tag{5}
\end{align}
\begin{equation}
	\label{eq6}
	\mathbf{Q}_{h}(d+1) =\mathbf{Q}_{h}(d)-\eta \frac{\partial H}{\partial \mathbf{Q}_{h}(d)}\tag{6}
\end{equation}where $\mathbf{z}^{\mathcal{S}}$ is student logits as shown in Fig.~\ref{fig_2}, $\mathbf{Q}_{h}$ is the consequent parameters of student model, $H$ is the cross-entropy loss, $\eta$ is the given learning rate.
\subsection{Least Learning Machine based Decoupling Knowledge Distillation (LLM-DKD)}
\subsubsection{Teacher Logits}
LLM-DKD takes the Negative Euclidean distance \cite{ref39} between teacher model output and each class as teacher logits:
\begin{equation}
	\label{eq7}
	z_{t}^{\mathcal{M}}=-\sqrt{\left(y^{\mathcal{M}}-\hat{y}_{t}\right)^{2}}\tag{7}
\end{equation}where $\hat{y}_{t}$ is the label of the $t$-th class, $\mathbf{z}^{\mathcal{M}}$ is teacher logits as shown in Fig.~\ref{fig_2}, $\mathbf{z}^{\mathcal{M}}=\left[z^{\mathcal{M}}_{1},z^{\mathcal{M}}_{2},\ldots,z^{\mathcal{M}}_{t},\ldots,z^{\mathcal{M}}_{C}\right]\in \mathrm{R}^{1\times C}$, $C$ is the number of classes.

\subsubsection{Target Class Knowledge and Non-target Class Knowledge}
For a given data from the $t$-th class, the soft labels can be denoted as $\mathbf{u}=[u_{1},u_{2},\ldots,u_{t},\ldots,u_{C}]\in \mathrm{R}^{1\times C}$, where $u_{i}$ is the soft label of the $i$-th class. Each element in $\mathbf{u}$ can be obtained by the \emph{softmax} function with temperature $\tau$:
\begin{equation}
	\label{eq8a}
	u_{i}=\frac{\exp \left(z_{i} / \tau\right)}{\sum_{j=1}^{C} \exp \left(z_{j} / \tau\right)}\tag{8a}
\end{equation}where $z_{i}$ represents the logit of the $i$-th class.

LLM-DKD decouples fuzzy dark knowledge into target class knowledge and non-target class knowledge by:
\begin{align}
	\label{eq8b}
	u_{t}&=\frac{\exp \left(z_{t} / \tau\right)}{\sum_{j=1}^{C} \exp \left(z_{j} / \tau\right)} & u_{\backslash t}&=\frac{\sum_{j^{\prime}=1, j^{\prime} \neq t}^{C} \exp \left(z_{j^{\prime}} / \tau\right)}{\sum_{j=1}^{C} \exp \left(z_{j} / \tau\right)}\tag{8b}
\end{align}where $u_{t}$ represents soft labels of the target class, containing knowledge about the “difficulty” of data. $u_{\backslash t}$ represents soft labels of non-target class, containing knowledge making knowledge distillation work \cite{ref32}.

We use $\mathbf{\hat{u}}=\left[\hat{u}_{1},\hat{u}_{2},\ldots,\hat{u}_{t-1},\hat{u}_{t+1},\ldots,\hat{u}_{C}\right]\in \mathrm{R}^{1\times (C-1)}$ to independently model probabilities among non-target classes:
\begin{equation}
	\label{eq8c}
	\hat{u}_{i}=\frac{\exp \left( z_{i} / \tau \right)}{\sum_{j=1,j \neq t}^{C} \exp \left(z_{j} / \tau \right)}\tag{8c}
\end{equation}

\begin{algorithm}
	\caption{Teacher model and Student model.}  
	\label{alg:A}  
	{\bf Input:}
	Training dataset $\mathbf{X}=\{\mathbf{x}_{i},\mathbf{x}_{i} \in \mathrm{R}^{m},i=1,2,...,N\}$ and the ground-truth label $\overline{\mathbf{Y}}=\{\overline{Y}_{i},\overline{Y}_{i}\in \mathrm{R},i=1,2,...,N\}$, the number of fuzzy rules $K$, the regularization parameter $L$, the maximum iteration epoch $\theta$, the threshold parameter $\xi$, the learning rate $\eta$.
	\\
	{\bf Output:} 
	the outputs of teacher model and student model. 
	\\
	{\bf Procedure:}
	\\
	{\bf Step 1}: Randomly select the center $v_{i}^{k}$ of Gaussian membership function in (2b) from five fixed fuzzy partition $\{0, 0.25, 0.5, 0.75, 1\}$, set the width $\delta_{i}^{k}$ to be a positive value, and compute the normalized fuzzy membership degree by (\ref{eq2a})-(\ref{eq2b}).
	\\
	{\bf Step 2}: Calculate antecedent parameter matrixs of teacher model and student model by (\ref{eq3a})-(\ref{eq3c}).
	\\
	{\bf Step 3}: The consequent parameter of teacher model $\mathbf{q}_{g}$ can be determined by using
	$\mathbf{q}_{g} = \left((1 / L) \mathbf{I}+\mathbf{X}_{g}^{T} \mathbf{X}_{g}\right)^{-1} \mathbf{X}_{g}^{T} \overline{\mathbf{Y}}$.
	\\
	{\bf Step 4}: The consequent parameter of student model $\mathbf{Q}_{h}$ can be updated using the gradient descent approach:
	\\
	\hspace*{0.2in}{\bf Step 4(a)}: Initialize consequent parameter $\mathbf{Q}_{h}$ and $d=1$.
	\\
	\hspace*{0.2in}Repeat
	\\
	\hspace*{0.2in}{\bf Step 4(b)}: Use (\ref{eq6}) to compute $\mathbf{Q}_{h}(d+1)$.
	\\
	\hspace*{0.2in}{\bf Step 4(c)}: $d=d+1$.
	\\
	\hspace*{0.2in}Until $H(d)-H(d-1) \leq \xi$ or $d \geq \theta$
	\\
	{\bf Step 5}: Calculate the output of teacher model $y^{\mathcal{M}}=\mathbf{q}_{g}^{T}\mathbf{x}_{g}$ and the output of student model $\mathbf{z}^{\mathcal{S}}=\mathbf{Q}_{h}^{T}\mathbf{x}_{h}$.
\end{algorithm}

\begin{algorithm}
	\caption{HTSK-LLM-DKD.}  
	\label{alg:A}  
	{\bf Input:}
	Training dataset $\mathbf{X}=\{\mathbf{x}_{i},\mathbf{x}_{i} \in \mathrm{R}^{m},i=1,2,...,N\}$ and the ground-truth label $\overline{\mathbf{Y}}=\{\overline{Y}_{i},\overline{Y}_{i} \in \mathrm{R},i=1,2,...,N\}$, the outputs of teacher model $\mathbf{y}^{\mathcal{M}}=\{y^{\mathcal{M}}_{i},y^{\mathcal{M}}_{i} \in \mathrm{R},i=1,2,...,N\}$ and the outputs of student model $\mathbf{Z}^{\mathcal{S}}=\{\mathbf{z}^{\mathcal{S}}_{i},\mathbf{z}^{\mathcal{S}}_{i} \in \mathrm{R}^{C},i=1,2,...,N\}$, the maximum iteration epoch $\theta$, the threshold parameter $\xi$, the learning rate $\eta$, the distillation parameters $\tau$, $\zeta$, $\lambda$, $\varphi$.
	\\
	{\bf Output:} 
	the output of HTSK-LLM-DKD. 
	\\
	{\bf Procedure:}
	\\
	{\bf Step 1}: Calculate the logits of teacher model $\mathbf{z}^{\mathcal{M}}$ with the Negative Euclidean distance between $y^{\mathcal{M}}$ and the label of each class by (\ref{eq7}).
	\\
	{\bf Step 2}: Calculate soft labels of teacher model $\mathbf{u}^{\mathcal{M}}$ and student model $\mathbf{u}^{\mathcal{S}}$ with the \emph{softmax} function by (\ref{eq8a}).
	\\
	{\bf Step 3}: Decouple fuzzy dark knowledge into target class knowledge $\mathbf{r}$ and non-target class knowledge $\mathbf{\hat{u}}$ by (\ref{eq8a})-(\ref{eq8c}).
	\\
	{\bf Step 4}: Use $\mathbf{r}$ and $\mathbf{\hat{u}}$ to rephrase $\operatorname{KL}(\mathbf{u}^{\mathcal{M}} \| \mathbf{u}^{\mathcal{S}})$ by (\ref{eq9a})-(\ref{eq9e}).
	\\
	{\bf Step 5}: Calculate the new loss function of HTSK-LLM-DKD:
	$Loss  = \zeta\operatorname{KL}\left(\mathbf{r}^{\mathcal{M}} \| \mathbf{r}^{\mathcal{S}}\right)+\lambda\mathrm{KL}\left(\widehat{\mathbf{u}}^{\mathcal{M}} \| \widehat{\mathbf{u}}^{\mathcal{S}}\right)+\varphi H$.
	\\
	{\bf Step 6}: Calculate the consequent parameter $\mathbf{Q}_{h}$ of HTSK-LLM-DKD using the gradient descent approach with the new loss function:
	\\
	\hspace*{0.2in}{\bf Step 6(a)}: Initialize consequent parameter $\mathbf{Q}_{h}$ and $d=1$.
	\\
	\hspace*{0.2in}Repeat
	\\
	\hspace*{0.2in}{\bf Step 6(b)}: $\mathbf{Q}_{h}(d+1) =\mathbf{Q}_{h}(d)-\eta \frac{\partial Loss}{\partial \mathbf{Q}_{h}(d)}$.
	\\
	\hspace*{0.2in}{\bf Step 6(c)}: $d=d+1$.
	\\
	\hspace*{0.2in}Until $Loss(d)-Loss(d-1) \leq \xi$ or $d \geq \theta$
	\\
	{\bf Step 7}: Calculate the output of HTSK-LLM-DKD.
\end{algorithm}

\subsubsection{Fuzzy Dark Knowledge Decoupling Process of LLM-DKD}
Widely used KL divergence \cite{ref14} is employed for decoupling fuzzy dark knowledge, which can be expressed as:
\begin{align*}
	\label{eq9a}
	KD=&\operatorname{KL}\left(\mathbf{u}^{\mathcal{M}} \| \mathbf{u}^{\mathcal{S}}\right)=u_{t}^{\mathcal{M}} \log \left(\frac{u_{t}^{\mathcal{M}}}{u_{t}^{\mathcal{S}}}\right)\\
	&+\sum_{i=1, i \neq t}^{C} u_{i}^{\mathcal{M}} \log \left(\frac{u_{i}^{\mathcal{M}}}{u_{i}^{\mathcal{S}}}\right)\tag{9a}
\end{align*}According to (\ref{eq8a})-(\ref{eq8c}), we can obtain that $\hat{u}_{i}={u}_{i}/u_{\backslash t}$, and (\ref{eq9a}) can be reformulated as:
\begin{align*}
	\label{eq9b}
	KD = &u_{t}^{\mathcal{M}} \log \left(\frac{u_{t}^{\mathcal{M}}}{u_{t}^{\mathcal{S}}}\right)\\
	&+u_{\backslash t}^{\mathcal{M}} \sum_{i=1, i \neq t}^{C} \hat{u}_{i}^{\mathcal{M}}\left(\log \left(\frac{\hat{u}_{i}^{\mathcal{M}}}{\hat{u}_{i}^{\mathcal{S}}}\right)+\log \left(\frac{u_{\backslash t}^{\mathcal{M}}}{u_{\backslash t}^{\mathcal{S}}}\right)\right)\tag{9b}
\end{align*}Since $u_{\backslash t}^{\mathcal{M}}$ and $u_{\backslash t}^{\mathcal{S}}$
are irrelevant to the class index $i$, (\ref{eq9b}) can be further expressed as:
\begin{align*}
	\label{eq9c}
	KD=&u_{t}^{\mathcal{M}} \log \left(\frac{u_{t}^{\mathcal{M}}}{u_{t}^{\mathcal{S}}}\right)+u_{\backslash t}^{\mathcal{M}} \log \left(\frac{u_{\backslash t}^{\mathcal{M}}}{u_{\backslash t}^{\mathcal{S}}}\right)\\
	&+u_{\backslash t}^{\mathcal{M}} \sum_{i=1, i \neq t}^{C} \hat{u}_{i}^{\mathcal{M}} \log \left(\frac{\hat{u}_{i}^{\mathcal{M}}}{\hat{u}_{i}^{\mathcal{S}}}\right)\tag{9c}
\end{align*}We define binary prediction $\mathbf{r}=[u_{t},u_{\backslash t}]\in \mathrm{R}^{1\times 2}$ to represent soft labels of target class and non-target class. The new expression of knowledge distillation can be described as:
\begin{equation}
	\label{eq9d}
	KD = \operatorname{KL}\left(\mathbf{r}^{\mathcal{M}} \| \mathbf{r}^{\mathcal{S}}\right)+\left(1-u_{t}^{\mathcal{M}}\right) \mathrm{KL}\left(\widehat{\mathbf{u}}^{\mathcal{M}} \| \widehat{\mathbf{u}}^{\mathcal{S}}\right)\tag{9d}
\end{equation}where $\mathbf{r}^{\mathcal{M}}$ and $\mathbf{r}^{\mathcal{S}}$ represent the soft labels of teacher model and student model, respectively.

As shown in (\ref{eq9d}), the loss function of knowledge distillation is reformulated as the weighted summation of two terms. $\operatorname{KL}\left(\mathbf{r}^{\mathcal{M}} \| \mathbf{r}^{\mathcal{S}}\right)$ indicates the similarity of binary probability for teacher and student model about target class prediction, called target class knowledge; $\mathrm{KL}\left(\widehat{\mathbf{u}}^{\mathcal{M}} \| \widehat{\mathbf{u}}^{\mathcal{S}}\right)$ indicates the similarity of teacher and student model for internal relations in non-target class, called non-target class knowledge. The transfer efficiency of non-target knowledge is negatively correlated with $u_{t}^{\mathcal{M}}$. We introduce $\zeta$ and $\lambda$ to reweight them, and get the following expression of decoupled knowledge distillation:
\begin{equation}
	\label{eq9e}
	DKD  = \zeta\operatorname{KL}\left(\mathbf{r}^{\mathcal{M}} \| \mathbf{r}^{\mathcal{S}}\right)+\lambda \mathrm{KL}\left(\widehat{\mathbf{u}}^{\mathcal{M}} \| \widehat{\mathbf{u}}^{\mathcal{S}}\right)\tag{9e}
\end{equation}
We integrate the cross-entropy loss $H$ in (\ref{eq5}) to obtain the total loss function of HTSK-LLM-DKD:
\begin{equation}
	\label{eq10}
	Loss  = \zeta\operatorname{KL}\left(\mathbf{r}^{\mathcal{M}} \| \mathbf{r}^{\mathcal{S}}\right)+\lambda\mathrm{KL}\left(\widehat{\mathbf{u}}^{\mathcal{M}} \| \widehat{\mathbf{u}}^{\mathcal{S}}\right)+\varphi H\tag{10}
\end{equation}
\subsection{HTSK-LLM-DKD Algorithm}
Here, we summarize the learning algorithm of teacher model and student model in Algorithm 1. The proposed HTSK-LLM-DKD is given in Algorithm 2.

\section{Experiments}
Our HTSK-LLM-DKD is an interpretable model, for its interpretability guarantee, nine fuzzy classifiers are selected to do comparative experiments with HTSK-LLM-DKD. In Section \uppercase\expandafter{\romannumeral4}-A, we describe the experiment setups. In Section \uppercase\expandafter{\romannumeral4}-B, we report the experimental results and analysis on UCI datasets. In Section \uppercase\expandafter{\romannumeral4}-C, a case study of HTSK-LLM-DKD on the \textit{Cleveland heart disease} dataset is given in detail. Section \uppercase\expandafter{\romannumeral4}-D discusses the effectiveness of decoupled knowledge distillation. In Section \uppercase\expandafter{\romannumeral4}-E, we explain the interpretability of HTSK-LLM-DKD.
\subsection{Experiment Setups and Performance Indicators}
\subsubsection{Datasets}
Since we focus on classification tasks, in this experiment, we randomly select sixteen widely used classification datasets from UCI repository \cite{ref35} and a real dataset \textit{Cleveland heart disease}. All the adopted dataset are normalized and the categorical features are converted into numerical features. Table S2 discribes all the adopted UCI datasets in the supplementary file.
\subsubsection{Comparative Methods}
As stated in Section III, HTSK-LLM-DKD is invented as a novel TSK fuzzy classifier by distilling fuzzy dark knowledge from High-order TSK fuzzy classifier to Low-order TSK fuzzy classifier. Therefore, we adopt $n$-order TSK fuzzy classifier ($n=0,1,2,3$) with several variants as comparative methods, \emph{i.e.}, TSK$_{v1}^{n}$ ($n$ is order, $n=0,1,2,3$) use LLM to solve consequent part; TSK$_{v2}^{n}$ ($n$ is order, $n=0,1,2,3$) take gradient descent approach to update consequent parameters; LSSVFS$^n$ ($n$ is order, $n=3$) utilizes the Least Squares Support Vector Machine (LSSVM) \cite{ref36} with third-order polynomial to solve consequent part. 
On the other hand, we realize three distillation models with two versions \emph{i.e}., CNN-TSK-KD and CNN-TSK-DKD, which all take CNN as teacher model, and First-order TSK fuzzy classifier as student model, employing traditional distillation method (denoted as KD) and the decoupled knowledge distillation method proposed in this paper (denoted as DKD), respectively. In addition, in oder to futher evalute the effectiveness of decoupled knowledge distillation method in HTSK-LLM-DKD, we implement HTSK-LLM-KD as comparative method, which use traditional distillation method in HTSK-LLM-DKD. Table S3 discribes all comparative methods in the supplementary file.
\subsubsection{Parameters setting}
All methods are cross-validated by ten-fold on the adopted datasets, and all adjustable parameters are optimized using grid search strategy. The optimization range of fuzzy rules is searched from $K=\lbrace1,2,...,20\rbrace$; the regularization parameter $L$ is set to $100$; distillation parameters $\tau$, $\zeta$, $\lambda$ and $\varphi$ are searched from $\lbrace1,2,5,10,20,100\rbrace$; the maximum iteration $\theta$ is set to 30; the threshold parameter $\xi$ is set to $10^{-5}$; the learning rate $\eta$ is set to 0.01; other parameters are set by the default values.
\subsubsection{Performance Indicators}
To evaluate the classification performance of all the adopted methods, four commonly-used performance indices are used in the experiment, \emph{i.e.}, $Accuracy$ (Acc for simplication), $Weighted-F$ (W-F for simplication), the running time and the average number of fuzzy rule. $Accuracy$ is the most intuitive performance metric, which displays the ratio of properly predicted samples to total samples. $Weighted-F$ is a weighted average of $Precision$ and $Recall$, which takes into account both false positives and false negatives. The best results are marked in bold, please note that ‘-’ means that the adopted methods cannot work out its results within 3 hours.

\subsection{Experimental Results and Analysis on UCI Datasets}
Table~\ref{table1} demonstrates the experiment results of the proposed HTSK-LLM-DKD and HTSK-LLM-KD compared with nine adopted fuzzy classifiers. We conclude the following results:
\begin{enumerate}
	\item{Our HTSK-LLM-DKD and HTSK-LLM-KD all achieve the first and the second best performance on eleven out of sixteen datasets  in terms of $Accuracy$ and $Weighted-F$, respectively, which shows that knowledge distillation can improve the performance of TSK fuzzy classifier very well and HTSK-LLM-DKD is much better than HTSK-LLM-KD. Additionally, HTSK-LLM-DKD keeps a comparable performance on \textit{TIT}, \textit{WIL}, \textit{PHO}, \textit{MAG} and \textit{ADU} datasets.}
	\item{Evidently, HTSK-LLM-DKD performs better with fewer fuzzy rules than TSK$_{v2}^{1}$, which is the student model of HTSK-LLM-DKD. At the same time, HTSK-LLM-DKD performs better with shorter running time than High-order TSK fuzzy classifier on most datasets. It shows that the fuzzy dark knowledge improves the generalization ability of HTSK-LLM-DKD greatly, and the proposed LLM-DKD can speed up the running time of HTSK-LLM-DKD.}
\end{enumerate}

Table~\ref{table2} displays the experiment results of HTSK-LLM-DKD, HTSK-LLM-KD, CNN-TSK-DKD and CNN-TSK-KD compared with its corresponding student model in terms of $Accuracy$ and $Weighted-F$, respectively. We conclude the following results:
\begin{enumerate}
	\item{Our HTSK-LLM-DKD has achieved the best performance improvement, with 1.41\% and 1.56\% in terms of $\emph{Accuracy}$ and $\emph{Weighted-F}$, which indicates that knowledge distillation can effectively improve the performance of student model by transferring fuzzy dark knowledge from teacher model.}
	\item{The comparative experiments in Table~\ref{table2} can be divided into two parts. The first part is the comparison between decoupled knowledge distillation and traditional knowledge distillation. It can be seen that HTSK-LLM-DKD and CNN-TSK-DKD perform better than HTSK-LLM-KD and CNN-TSK-KD, we believe the reason lie in that: in traditional knowledge distillation, the transfer efficiency of non-target knowledge is negatively correlated with the confidence on the target class of teacher model, and decoupled knowledge distillation can better transfer non-target knowledge by re-weighting, thus improving the generalization ability of student model.}
	\item{The second part is the comparison between CNN and TSK fuzzy classifier as teacher models. We can observe that the average $Accuracy$ and $Weighted-F$ of HTSK-LLM-KD and HTSK-LLM-DKD are usually higher than CNN-TSK-KD and CNN-TSK-DKD, which indicates that the fuzzy dark knowledge transfered from High-order TSK fuzzy classifier is more conducive to improving the performance of low-order TSK fuzzy classifier than CNN.}
\end{enumerate}


\begin{table*}[!t]
	\caption{Rule Number, Average Time, Average \emph{Accuracy} (\%), Average \emph{Weighted-F} (\%) and Standard Deviation (\%) on UCI Datasets}
	\label{table1}
	\centering
	\footnotesize
	\tabcolsep=1.5pt
	\begin{tabular}{|c|c|c|c|c|c|c|c|c|c|c|c|}
		\hline
		\multicolumn{1}{|c|}{\multirow{4}[4]{*}{Datasets}} & \makecell*[c]{TSK$_{v1}^{0}$} & TSK$_{v1}^{1}$ & TSK$_{v1}^{2}$ & TSK$_{v1}^{3}$ & TSK$_{v2}^{0}$ & TSK$_{v2}^{1}$ & TSK$_{v2}^{2}$ & TSK$_{v2}^{3}$ & LSSVFS$^3$ & \makecell[c]{HTSK\\-LLM\\-KD} & \makecell[c]{HTSK\\-LLM\\-DKD} \\
		\cline{2-12}          & Acc±Std & Acc±Std & Acc±Std & Acc±Std & Acc±Std & Acc±Std & Acc±Std & Acc±Std & Acc±Std & Acc±Std & Acc±Std \bigstrut[t]\\
		& W-F±Std & W-F±Std & W-F±Std & W-F±Std & W-F±Std & W-F±Std & W-F±Std & W-F±Std & W-F±Std & W-F±Std & W-F±Std \\
		& Rules   Time & Rules   Time & Rules   Time & Rules   Time & Rules   Time & Rules   Time & Rules   Time & Rules   Time & Rules   Time & Rules   Time & Rules   Time \\
		\hline
		\multicolumn{1}{|c|}{\multirow{3}[2]{*}{\textit{IRI}}} & 82.59±4.05 & 85.33±4.23 & 83.33±4.67 & 80.06±4.89 & 90.44±4.72 & 97.08±3.46 & \multicolumn{1}{c|}{96.80±3.49} & \multicolumn{1}{c|}{96.33±4.68} & \multicolumn{1}{c|}{85.40±4.25} & 98.00±3.22 & \textbf{98.66±2.81} \bigstrut[t]\\
		& 83.28±4.04 & 85.48±4.93 & 83.87±4.95 & 81.12±4.74 & 90.31±4.94 & 96.96±3.70 & \multicolumn{1}{c|}{96.39±4.07} & \multicolumn{1}{c|}{95.84±5.42} & \multicolumn{1}{c|}{86.15±4.52} & 98.06±3.16 & \textbf{98.56±3.01} \\
		& 18.1    \textbf{0.0050} & 10.3  0.0063 & 2.6    0.0087 & \textbf{1.1}    0.0133 & 17.4    0.1806 & 8.5    0.1421 & \multicolumn{1}{c|}{6.2    0.1583} & \multicolumn{1}{c|}{2.8    0.1825} & \multicolumn{1}{c|}{2.4   0.5021} & 7.9    0.1462 & 6.3    0.1511 \bigstrut[b]\\
		\hline
		\multicolumn{1}{|c|}{\multirow{3}[2]{*}{\textit{WIN}}} & 80.48±4.98 & 88.02±4.10 & 83.43±5.47 & 75.71±4.94 & 93.32±3.69 & 98.39±2.66 & \multicolumn{1}{c|}{97.20±3.07} & \multicolumn{1}{c|}{96.27±3.70} & \multicolumn{1}{c|}{89.50±3.53} & 98.88±2.34 & \textbf{99.44±1.75} \bigstrut[t]\\
		& 80.52±4.91 & 88.04±4.06 & 83.40±5.08 & 75.45±4.91 & 93.36±3.69 & 98.08±3.29 & \multicolumn{1}{c|}{97.01±3.39} & \multicolumn{1}{c|}{96.30±3.67} & \multicolumn{1}{c|}{89.49±3.61} & 98.48±3.22 & \textbf{99.15±2.67} \\
		& 18.2    \textbf{0.0049} & \textbf{4.4}   0.0130 & 8.1    0.1034 & 7.8    0.5818 & 18.3    0.1656 & 9.4    0.1622 & \multicolumn{1}{c|}{8.2    0.3519} & \multicolumn{1}{c|}{4.6   1.6877} & \multicolumn{1}{c|}{5   0.7188} & 8.5   0.1698 & 8.1   0.1532 \bigstrut[b]\\
		\hline
		\multicolumn{1}{|c|}{\multirow{3}[2]{*}{\textit{TIT}}} & 79.05±3.85 & 79.08±3.82 & 78.87±4.02 & \textbf{79.12±3.75} & 77.82±3.62 & 77.90±3.75 & \multicolumn{1}{c|}{78.82±3.69} & \multicolumn{1}{c|}{79.00±3.85} & \multicolumn{1}{c|}{78.46±3.99} & 78.41±4.09 & 78.60±3.87 \bigstrut[t]\\
		& 76.00±4.85 & 76.02±4.72 & 75.71±4.89 & \textbf{76.21±4.67} & 75.20±4.44 & 75.42±4.25 & \multicolumn{1}{c|}{75.42±4.02} & \multicolumn{1}{c|}{75.35±4.42} & \multicolumn{1}{c|}{75.39±4.38} & 75.34±4.91 & 75.41±4.50 \\
		& 16.3   \textbf{0.0103} & 6.2   0.0118 & 4.2    0.0135 & 3.2   0.0424 & 14.7   0.8405 & 7.5   0.8386 & \multicolumn{1}{c|}{6.8    0.9367} & \multicolumn{1}{c|}{5.9    1.1095} & \multicolumn{1}{c|}{\textbf{2.4}    93.4825} & 5.7    1.1072 & 4.8   0.9931 \bigstrut[b]\\
		\hline
		\multicolumn{1}{|c|}{\multirow{3}[2]{*}{\textit{SEE}}} & 92.04±5.30 & 92.90±5.10 & 91.90±5.04 & 82.85±5.75 & 92.32±4.56 & 92.50±4.64 & \multicolumn{1}{c|}{93.76±4.95} & \multicolumn{1}{c|}{91.66±5.76}& \multicolumn{1}{c|}{94.00±4.12} & 94.85±4.61 & \textbf{96.19±3.85} \bigstrut[t]\\
		& 92.05±5.41 & 92.91±5.12 & 91.92±5.09 & 82.96±5.69 & 92.30±4.66 & 92.19±4.98 & \multicolumn{1}{c|}{93.66±4.19} & \multicolumn{1}{c|}{91.47±5.23} & \multicolumn{1}{c|}{94.03±4.09} & 94.32±4.37 & \textbf{96.06±3.97} \\
		& 17.5    \textbf{0.0039} & 6.9   0.0112 & \textbf{1}   0.0100 & \textbf{1}    0.0594 & 19.1    0.1724 & 8.4    0.1733 & \multicolumn{1}{c|}{6.5    0.2075} & \multicolumn{1}{c|}{3.8    0.3060} & \multicolumn{1}{c|}{2.2   1.1360} & 8.9    0.2588 & 6.2    0.2152 \bigstrut[b]\\
		\hline
		\multicolumn{1}{|c|}{\multirow{3}[2]{*}{\textit{ION}}} & 82.59±5.99 & 86.43±5.06 & 80.57±5.31 & 79.34±5.34 & 81.91±5.33 & 92.40±4.77 & \multicolumn{1}{c|}{90.53±4.68} & \multicolumn{1}{c|}{90.10±5.36} & \multicolumn{1}{c|}{80.88±5.11} & 92.53±5.12 & \textbf{92.87±4.31} \bigstrut[t]\\
		& 81.75±5.70 & 85.76±5.49 & 79.49±5.25 & 78.13±5.29 & 81.12±5.42 & 91.44±5.26 & \multicolumn{1}{c|}{89.32±5.17} & \multicolumn{1}{c|}{88.97±5.15} & \multicolumn{1}{c|}{79.48±5.17} & 91.58±4.74 & \textbf{92.11±4.40} \\
		& 17.6    \textbf{0.0152} & \textbf{2.5}    0.0339 & 7.1    0.7432 & 6.5    20.3081 & 17.3   0.3414 & 13.4    0.3944 & \multicolumn{1}{c|}{6.9    2.8994} & \multicolumn{1}{c|}{4.6    44.3618} & \multicolumn{1}{c|}{5.2   2.7954} & 8.8    0.3069 & 7.3    0.3158 \bigstrut[b]\\
		\hline
		\multicolumn{1}{|c|}{\multirow{3}[2]{*}{\textit{WIL}}} & 95.05±1.09 & 97.89±0.81 & 97.54±0.94 & \textbf{97.99±0.79} & 94.60±1.03 & 95.36±0.95 & \multicolumn{1}{c|}{96.94±1.52} & \multicolumn{1}{c|}{97.18±1.11} & \multicolumn{1}{c|}{97.46±0.70} & 95.44±1.11 & 95.57±0.98 \bigstrut[t]\\
		& 94.10±1.77 & 97.74±0.93 & 97.44±1.06 & \textbf{97.83±0.88} & 93.98±1.51 & 94.45±1.56 & \multicolumn{1}{c|}{96.36±2.02} & \multicolumn{1}{c|}{96.72±2.15} & \multicolumn{1}{c|}{97.18±0.79} & 94.04±1.73 & 94.16±1.40 \\
		& 19.1    \textbf{0.0231} & 18.4    0.0758 & 8.8    0.1498 & 4.1    0.5598 & 17.3    1.5435 & 10.1   2.2452 & \multicolumn{1}{c|}{8.7    2.8232} & \multicolumn{1}{c|}{7.7    5.6222} & \multicolumn{1}{c|}{\textbf{1.6}  438.6728} & 9.4    3.0981 & 9.3    3.6865 \bigstrut[b]\\
		\hline
		\multicolumn{1}{|c|}{\multirow{3}[2]{*}{\textit{WIS}}} & 96.87±1.64 & 96.89±1.96 & 96.77±2.05 & 89.60±2.21 & 96.44±1.64 & 96.99±1.51 & \multicolumn{1}{c|}{96.55±1.88} & \multicolumn{1}{c|}{96.47±2.04} & \multicolumn{1}{c|}{95.81±1.65} & 97.07±1.38 & \textbf{97.80±1.24} \bigstrut[t]\\
		& 96.87±1.65 & 96.89±1.97 & 96.78±2.06 & 89.28±2.48 & 96.43±1.65 & 96.64±1.61 & \multicolumn{1}{c|}{96.18±2.07} & \multicolumn{1}{c|}{96.08±2.25} & \multicolumn{1}{c|}{95.77±1.67} & 97.01±1.45 & \textbf{97.52±1.40} \\
		& 19.3   \textbf{0.0235} & 7.3   0.0475 & \textbf{1}    0.0539 & \textbf{1}   0.4966 & 18.7   0.3595 & 10.7    0.3736 & \multicolumn{1}{c|}{7.8    0.6556} & \multicolumn{1}{c|}{2.9    1.4914} & \multicolumn{1}{c|}{9.1     10.8646} & 7.6    0.3545 & 6.1   0.3974 \bigstrut[b]\\
		\hline
		\multicolumn{1}{|c|}{\multirow{3}[2]{*}{\textit{QSA}}} & 77.80±4.88 & 86.21±3.19 & 79.30±4.30 & 78.57±5.57 & 77.88±4.42 & 88.02±2.62 & \multicolumn{1}{c|}{86.72±2.59} & \multicolumn{1}{c|}{87.00±3.19}  & \multicolumn{1}{c|}{79.15±4.72} & 88.03±2.06 & \textbf{88.05±3.33} \bigstrut[t]\\
		& 77.00±5.29 & 86.15±3.23 & 79.19±4.06 & 78.56±5.43 & 77.19±4.81 & 86.51±2.66 & \multicolumn{1}{c|}{85.05±2.69} & \multicolumn{1}{c|}{85.36±3.33} & \multicolumn{1}{c|}{77.39±5.23} & 86.51±2.08 & \textbf{86.53±3.41} \\
		& 19.3    \textbf{0.0179} & 7.4    0.3647 & 6.7    6.7294 & \textbf{1}   20.6744 & 19.2    0.8045 & 14.6    1.3554 & \multicolumn{1}{c|}{8.8    14.9634} & \multicolumn{1}{c|}{8.3    533.1391} & \multicolumn{1}{c|}{6.5   22.6447} & 10.6    0.9893 & 8.7    0.7985 \bigstrut[b]\\
		\hline
		\multicolumn{1}{|c|}{\multirow{3}[2]{*}{\textit{PHO}}} & 77.11±1.38 & 84.36±1.56 & 85.58±1.71 & \textbf{87.26±1.75} & 74.68±1.68 & 80.79±1.33 & \multicolumn{1}{c|}{83.38±1.64} & \multicolumn{1}{c|}{83.14±1.42} & \multicolumn{1}{c|}{84.62±1.91} & 80.45±1.12 & 81.51±1.20 \bigstrut[t]\\
		& 76.34±1.48 & 84.32±1.55 & 85.51±1.71 & \textbf{87.17±1.78} & 72.46±2.30 & 80.39±1.23 & \multicolumn{1}{c|}{83.00±2.26} & \multicolumn{1}{c|}{82.94±2.15} & \multicolumn{1}{c|}{84.11±2.49} & 80.27±1.86 & 81.23±1.71 \\
		& 18.5   \textbf{0.0267} & 17.9   0.0887 & 9.6    0.1892 & 9.2   3.3785 & 16.1   2.2788 & 15.1    2.9822 & \multicolumn{1}{c|}{9.3    3.2500} & \multicolumn{1}{c|}{6.5    5.4895} & \multicolumn{1}{c|}{\textbf{1.3}    443.6744} & 14.3   2.4684 & 14.5   2.6239 \bigstrut[b]\\
		\hline
		\multicolumn{1}{|c|}{\multirow{3}[2]{*}{\textit{SON}}} & 68.39±4.15 & 78.63±3.44 & 79.56±3.75 & 79.36±3.94 & 69.40±3.82 & 85.66±2.92 & \multicolumn{1}{c|}{84.55±1.88} & \multicolumn{1}{c|}{87.08±1.97} & \multicolumn{1}{c|}{81.79±3.18} & 87.15±2.41 & \textbf{88.02±2.13} \bigstrut[t]\\
		& 68.42±4.18 & 78.59±3.42 & 79.32±3.98 & 78.56±3.77 & 69.58±3.93 & 84.74±3.24 & \multicolumn{1}{c|}{83.30±2.55} & \multicolumn{1}{c|}{86.09±1.53} & \multicolumn{1}{c|}{81.65±2.24} & 86.11±2.36 & \textbf{86.88±2.35} \\
		& 14.4   \textbf{0.0114} & 13.8   0.1089 & 6.9    0.9521 & 5    42.6555 & 16.6    0.2687 & 10.2    0.2974 & \multicolumn{1}{c|}{6.8    4.8772} & \multicolumn{1}{c|}{5.5    208.4717} & \multicolumn{1}{c|}{\textbf{2.8}   0.9349} & 9.8    0.3065 & 9.3   0.2782 \bigstrut[b]\\
		\hline
		\multicolumn{1}{|c|}{\multirow{3}[2]{*}{\textit{SEG}}} & 92.49±3.90 & 99.54±0.50 & 99.43±0.50 & 98.53±0.81 & 86.51±2.67 & 99.61±0.43 & \multicolumn{1}{c|}{99.52±0.43} & \multicolumn{1}{c|}{99.56±0.53} & \multicolumn{1}{c|}{99.35±0.42} & 99.64±0.50 & \textbf{99.69±0.45} \bigstrut[t]\\
		& 91.20±3.41 & 99.52±0.46 & 99.40±0.58 & 98.66±0.74 & 84.31±2.61 & 99.13±0.97 & \multicolumn{1}{c|}{99.01±0.88} & \multicolumn{1}{c|}{99.11±1.10} & \multicolumn{1}{c|}{98.65±0.88} & 99.19±1.03 & \textbf{99.40±0.86} \\
		& 18.6   \textbf{0.0198} & 9.2   0.0610 & \textbf{1}    0.1318 & \textbf{1}    9.7983 & 16.5    1.1173 & 16.2    1.8406 & \multicolumn{1}{c|}{8.1    5.3774} & \multicolumn{1}{c|}{7.8    78.6618} & \multicolumn{1}{c|}{7.9   107.0428} & 13.6    1.5924 & 8.1   1.1664 \bigstrut[b]\\
		\hline
		\multicolumn{1}{|c|}{\multirow{3}[2]{*}{\textit{VOT}}} & 90.89±4.88 & 94.02±3.64 & 93.53±4.23 & 93.22±3.82 & 90.42±4.00 & 94.81±3.61 & \multicolumn{1}{c|}{93.53±3.41} & \multicolumn{1}{c|}{93.82±3.58} & \multicolumn{1}{c|}{94.38±3.18} & 94.93±3.04 & \textbf{96.07±3.09} \bigstrut[t]\\
		& 90.93±4.83 & 94.06±3.61 & 93.55±4.20 & 93.18±3.85 & 90.44±4.00 & 94.83±3.86 & \multicolumn{1}{c|}{93.81±3.85} & \multicolumn{1}{c|}{93.01±4.08} & \multicolumn{1}{c|}{93.91±3.16} & 94.90±3.31 & \textbf{95.72±3.44} \\
		& 18.1   \textbf{0.0142} & 6.5    0.0558 & \textbf{1}   0.1160 & 3.1   1.2698 & 18.2   0.2776 & 9.6   0.2836 & \multicolumn{1}{c|}{8.4    0.8319} & \multicolumn{1}{c|}{4.9    7.0024} & \multicolumn{1}{c|}{7.1     4.1051} & 4.1    0.3036 & 5.4   0.3140 \bigstrut[b]\\
		\hline
		\multicolumn{1}{|c|}{\multirow{3}[2]{*}{\textit{CAR}}} & 79.35±3.12 & 88.67±2.64 & 88.13±2.72 & 88.80±2.51 & 82.61±2.35 & 90.84±2.20 & \multicolumn{1}{c|}{90.75±2.22} & \multicolumn{1}{c|}{90.89±2.34} & \multicolumn{1}{c|}{89.79±1.41} & 90.92±2.47 & \textbf{90.96±1.83} \bigstrut[t]\\
		& 78.79±3.16 & 88.86±2.54 & 88.28±2.58 & 88.99±2.37 & 81.08±3.61 & 90.33±2.55 & \multicolumn{1}{c|}{90.79±2.04} & \multicolumn{1}{c|}{90.84±2.41} & \multicolumn{1}{c|}{89.23±1.52} & 90.65±2.51 & \textbf{90.66±1.93} \\
		& 17.7   \textbf{0.0207} & 16.6   0.1723 & 2.1    0.9202 & \textbf{1}   12.3888 & 17.9   1.1823 & 16.5    1.9752 & \multicolumn{1}{c|}{8.1    6.4973} & \multicolumn{1}{c|}{7.9    113.7304} & \multicolumn{1}{c|}{8.2    87.4473} & 12.4    1.7951 & 9.3   1.3367 \bigstrut[b]\\
		\hline
		\multicolumn{1}{|c|}{\multirow{3}[2]{*}{\textit{MAG}}} & 79.48±1.23 & 86.39±0.72 & 86.23±1.31 & \textbf{86.94±1.63} & 76.65±1.98 & 85.39±0.99 & \multicolumn{1}{c|}{84.97±1.32} & \multicolumn{1}{c|}{82.86±0.52} & \multicolumn{1}{c|}{86.61±0.98} & 85.41±0.76 & 85.59±0.91 \bigstrut[t]\\
		& 78.75±1.75 & 85.32±0.53 & 84.82±1.64 & \textbf{85.88±1.36} & 75.87±2.63 & 84.57±1.02 & \multicolumn{1}{c|}{84.06±1.68} & \multicolumn{1}{c|}{81.74±0.86} & \multicolumn{1}{c|}{85.52±1.05} & 84.22±1.58 & 84.58±0.96 \\
		& 18.3    \textbf{0.1075} & 15.6    0.4689 & 6.9   34.7426 & 2.7    184.4161 & 18.6   8.9985 & 17.3    13.8653 & \multicolumn{1}{c|}{5.7    13.5862} & \multicolumn{1}{c|}{4.3    60.2334} & \multicolumn{1}{c|}{\textbf{1}   6870.9871} & 16.9   12.2797 & 16.3    12.8768 \bigstrut[b]\\
		\hline
		\multicolumn{1}{|c|}{\multirow{3}[2]{*}{\textit{BRA}}} & 94.92±0.74 & 95.27±0.41 & 95.38±0.37 & 95.43±0.47 & 95.17±0.48 & 95.54±0.58 & \multicolumn{1}{c|}{95.51±0.53} & \multicolumn{1}{c|}{95.55±0.67} & \multicolumn{1}{c|}{95.51±0.18} & 95.58±0.44 & \textbf{95.61±0.32} \bigstrut[t]\\
		& 94.85±0.68 & 95.23±0.48 & 95.28±0.38 & 95.33±0.45 & 95.04±0.54 & 95.48±0.54 & \multicolumn{1}{c|}{95.42±0.57} & \multicolumn{1}{c|}{95.50±0.62} & \multicolumn{1}{c|}{95.46±0.17} & 95.51±0.46 & \textbf{95.52±0.36} \\
		& 18.2   \textbf{0.0653} & 13.3   0.1503 & 5.6   0.6842 & \textbf{2.9}   0.7094 & 17.4   8.1934 & 10.7    8.5122 & \multicolumn{1}{c|}{4.3    8.8453} & \multicolumn{1}{c|}{3.7    9.7481} & \multicolumn{1}{c|}{6.8   8.8453} & 11.7    8.5681 & 11.1   8.9657 \bigstrut[b]\\
		\hline
		\multicolumn{1}{|c|}{\multirow{3}[2]{*}{\textit{ADU}}} & 80.32±1.57 & 84.55±0.78 & 84.95±0.68 & \textbf{85.28±0.84} & 78.99±1.54 & 84.36±0.84 & \multicolumn{1}{c|}{\multirow{3}[2]{*}{-}} & \multicolumn{1}{c|}{\multirow{3}[2]{*}{-}} & \multicolumn{1}{c|}{\multirow{3}[2]{*}{-}} & 84.44±0.83 & 84.46±0.87 \bigstrut[t]\\
		& 77.25±2.58 & 83.58±0.87 & 84.16±0.82 & \textbf{84.48±0.95} & 74.24±2.82 & 83.92±0.74 &       &       &       & 83.99±0.87 & 84.04±1.83 \\
		& 19.6   \textbf{0.2658} & 17.1   1.7375 & 7.2    7.8395 & \textbf{1.3}   55.6825 & 19.3   22.8843 & 17.5    36.8652 &       &       &       & 16.4    36.5892 & 16.3   37.9834 \bigstrut[b]\\
		\hline
	\end{tabular}
\end{table*}

\begin{table*}[htbp]
	\caption{Comparison of \emph{Accuracy} (\%) and \emph{Weighted-F} (\%) Improvements of HTSK-LLM-DKD and CNN-TSK-DKD Models on UCI Datasets}
	\label{table2}
	\centering
	\footnotesize
	\tabcolsep=1.5pt
	\begin{tabular}{|c|c|c|c|c|c|c|c|c|c|c|c|c|}
		\hline
		\multirow{3}[6]{*}{Datasets} & \multicolumn{3}{c|}{CNN-TSK-KD} & \multicolumn{3}{c|}{CNN-TSK-DKD} & \multicolumn{3}{c|}{HTSK-LLM-KD} & \multicolumn{3}{c|}{HTSK-LLM-DKD} \bigstrut[t]\\
		\cline{2-13}    \multicolumn{1}{|c|}{} & Distillation & Student & Promotion & Distillation & Student & Promotion & Distillation & Student & Promotion & Distillation & Student & Promotion \bigstrut[t]\\
		\cline{2-13}    \multicolumn{1}{|c|}{} & Acc/W-F & Acc/W-F & Acc/W-F & Acc/W-F & Acc/W-F & Acc/W-F & Acc/W-F & Acc/W-F & Acc/W-F & Acc/W-F & Acc/W-F & Acc/W-F \bigstrut[t]\\
		\hline
		\textit{IRI} & 97.66/97.24 & 95.99/95.14 & 1.67/2.10 & 98.33/98.06 & 95.66/94.80 & 2.67/3.26 & 98.00/98.06 & 95.96/95.71 & 2.04/2.35 & 98.66/98.56 & 95.66/94.98 & 3.00/3.58 \bigstrut[t]\\
		\hline
		\textit{WIN} & 99.18/98.85 & 97.62/97.21 & 1.56/1.64 & 99.24/98.93 & 97.54/97.15 & 1.70/1.78 & 98.88/98.48 & 97.73/97.31 & 1.15/1.17 & 99.44/99.15 & 97.63/97.33 & 1.81/1.82 \bigstrut[t]\\
		\hline
		\textit{TIT} & 78.23/75.20 & 77.41/74.12 & 0.82/1.08 & 78.28/75.10 & 77.28/73.51 & 1.00/1.59 & 78.41/75.34 & 77.50/73.84 & 0.91/1.50 & 78.60/75.41 & 77.60/73.82 & 1.00/1.59 \bigstrut[t]\\
		\hline
		\textit{SEE} & 93.90/93.79 & 92.00/91.42 & 1.90/2.37 & 95.76/95.61 & 91.00/90.50 & 4.76/5.11 & 94.85/94.32 & 91.99/91.10 & 2.86/3.22 & 96.19/96.06 & 91.42/91.02 & 4.77/5.04 \bigstrut[t]\\
		\hline
		\textit{ION} & 92.62/91.89 & 91.89/91.05 & 0.73/0.84 & 92.78/91.95 & 91.35/90.37 & 1.43/1.58 & 92.53/91.58 & 91.97/90.95 & 0.56/0.63 & 92.87/92.11 & 91.73/90.78 & 1.14/1.33 \bigstrut[t]\\
		\hline
		\textit{WIL} & 95.52/93.89 & 93.79/91.83 & 1.73/2.06 & 95.78/94.36 & 93.82/91.99 & 1.96/2.37 & 95.44/93.46 & 93.86/91.75 & 1.58/1.71 & 95.57/94.16 & 93.80/91.92 & 1.77/2.24 \bigstrut[t]\\
		\hline
		\textit{WIS} & 96.95/96.74 & 96.53/96.26 & 0.42/0.48 & 97.35/97.06 & 96.48/96.11 & 0.87/0.95 & 97.07/97.01 & 96.63/96.50 & 0.44/0.51 & 97.80/97.52 & 96.92/96.53 & 0.88/0.99 \bigstrut[t]\\
		\hline
		\textit{QSA} & 87.79/86.27 & 87.03/85.30 & 0.76/0.97 & 87.94/86.41 & 87.09/85.32 & 0.85/1.09 & 88.03/86.51 & 87.17/85.51 & 0.86/1.00 & 88.05/86.53 & 87.10/85.41 & 0.95/1.12 \bigstrut[t]\\
		\hline
		\textit{PHO} & 80.32/80.15 & 79.98/79.79 & 0.34/0.36 & 81.16/80.96 & 79.59/79.24 & 1.57/1.72 & 80.45/80.27 & 79.89/79.66 & 0.56/0.61 & 81.51/81.23 & 79.73/79.36 & 1.78/1.87 \bigstrut[t]\\
		\hline
		\textit{SON} & 87.23/86.33 & 85.18/84.24 & 2.05/2.09 & 88.08/86.97 & 85.13/83.99 & 2.95/2.98 & 87.15/86.11 & 85.19/84.23 & 1.96/1.88 & 88.02/86.88 & 85.12/83.96 & 2.90/2.92 \bigstrut[t]\\
		\hline
		\textit{SEG} & 99.63/99.17 & 99.56/99.05 & 0.07/0.12 & 99.67/99.37 & 99.37/98.96 & 0.30/0.41 & 99.64/99.19 & 99.55/99.04 & 0.09/0.15 & 99.69/99.40 & 99.30/98.92 & 0.39/0.48 \bigstrut[t]\\
		\hline
		\textit{VOT} & 94.85/94.65 & 94.45/94.44 & 0.40/0.21 & 95.89/95.54 & 94.59/94.52 & 1.30/1.02 & 94.93/94.90 & 94.41/94.67 & 0.52/0.23 & 96.07/95.72 & 94.54/94.52 & 1.53/1.20 \bigstrut[t]\\
		\hline
		\textit{CAR} & 90.88/90.56 & 90.66/90.26 & 0.22/0.30 & 90.93/90.66 & 90.65/90.26 & 0.28/0.40 & 90.92/90.65 & 90.68/90.31 & 0.24/0.34 & 90.96/90.66 & 90.68/90.30 & 0.28/0.36 \bigstrut[t]\\
		\hline
		\textit{MAG} & 84.75/82.22 & 84.68/82.11 & 0.07/0.11 & 85.50/84.46 & 85.31/84.25 & 0.19/0.21 & 85.41/84.22 & 85.36/84.15 & 0.05/0.07 & 85.59/84.58 & 85.39/84.37 & 0.20/0.21 \bigstrut[t]\\
		\hline
		\textit{BRA} & 95.56/95.49 & 95.49/95.42 & 0.07/0.07 & 95.59/95.52 & 95.51/95.44 & 0.08/0.08 & 95.58/95.51 & 95.51/95.43 & 0.07/0.08 & 95.61/95.52 & 95.53/95.44 & 0.08/0.08 \bigstrut[t]\\
		\hline
		\textit{ADU} & 84.39/83.90 & 84.30/83.81 & 0.09/0.09 & 84.43/83.96 & 84.31/83.83 & 0.12/0.13 & 84.44/83.99 & 84.33/83.87 & 0.11/0.12 & 84.46/84.04 & 84.31/83.87 & 0.15/0.17 \bigstrut[t]\\
		\hline
		Average & 91.21/90.39 & 90.41/89.46 & 0.80/0.93 & 91.66/90.93 & 90.29/89.39 & 1.37/1.54 & 91.35/90.60 & 90.48/89.62 & 0.87/0.97 & 91.81/91.09 & 90.40/89.53 & 1.41/1.56 \bigstrut[t]\\
		\hline
	\end{tabular}%
\end{table*}%

\begin{table}[htbp]
	\caption{Average \emph{Accuracy} (\%), \emph{Weighted-F} (\%) with Corresponding Standard Deviation (\%), Rule Number and Running Time on \emph{CLE}}
	\label{table3}
	\centering
	\footnotesize
	\tabcolsep=1.5pt
	\begin{tabular}{|c|c|c|c|c|}
		\hline
		Methods & \multicolumn{1}{c|}{Acc±Std} & \multicolumn{1}{c|}{W-F±Std} & Rules & Time \bigstrut[t]\\
		\hline
		TSK$_{v1}^{0}$ & \multicolumn{1}{c|}{70.28±2.65} & \multicolumn{1}{c|}{69.87±3.25} & 16.6    & \textbf{0.0079} \bigstrut[t]\\
		\hline
		TSK$_{v1}^{1}$ & 73.28±3.72 & 73.54±3.38 & 9.3     & 0.0472 \bigstrut[t]\\
		\hline
		TSK$_{v1}^{2}$ & 70.98±3.63 & 70.96±3.83 & \textbf{1} & 0.0168 \bigstrut[t]\\
		\hline
		TSK$_{v1}^{3}$ & 70.25±4.68 & 70.24±4.34 & 1.2    & 0.2073 \bigstrut[t]\\
		\hline
		TSK$_{v2}^{0}$ & 73.68±3.77 & 72.38±3.24 & 17.5    & 0.2475 \bigstrut[t]\\
		\hline
		TSK$_{v2}^{1}$ & 75.85±3.67 & 75.68±3.35 & 10.6     & 0.3275 \bigstrut[t]\\
		\hline
		TSK$_{v2}^{2}$ & 71.35±2.64 & 71.12±3.36 & 2.6    & 1.7883 \bigstrut[t]\\
		\hline
		TSK$_{v2}^{3}$ & 71.75±2.35 & 71.83±2.57 & \textbf{1}    & 1.0838 \bigstrut[t]\\
		\hline
		LSSVFS$^{3}$ & 69.68±4.35 & 69.38±4.79 & 6.6 & 2.1436 \bigstrut[t]\\
		\hline
		\makecell[c]{HTSK-LLM-KD} & 78.63±3.83 & 78.52±3.96 & 8.6     & 0.2782 \bigstrut[t]\\
		\hline
		\makecell[c]{HTSK-LLM-DKD} & \textbf{79.10±3.52} & \textbf{79.08±3.35} & 8.1     & 0.3168 \bigstrut[t]\\
		\hline
	\end{tabular}%
	\label{tab:addlabel}%
\end{table}%

\begin{table}[htbp]
	\caption{Comparison of The \emph{Accuracy} (\%) and \emph{Weighted-F} (\%) Improvement of HTSK-LLM-DKD and CNN-TSK-DKD Models on \emph{CLE}}
	\label{table4}
	\centering
	\footnotesize
	\tabcolsep=1.5pt
	\begin{tabular}{|c|c|c|c|}
		\hline
		\multicolumn{1}{|c|}{\multirow{2}[4]{*}{Methods}} & \multicolumn{3}{c|}{\emph{Accuracy}/\emph{Weighted-F}} \bigstrut[t]\\
		\cline{2-4}    \multicolumn{1}{|c|}{} & \multicolumn{1}{c|}{Distillation} & \multicolumn{1}{c|}{Student} & \multicolumn{1}{c|}{Promotion} \bigstrut[t]\\
		\hline
		CNN-TSK-KD & 78.56/78.50  & 75.19/75.15  & 3.37/3.35 \bigstrut[t]\\
		\hline
		CNN-TSK-DKD & 78.80/78.69  & 75.35/75.25  & 3.45/3.44 \bigstrut[t]\\
		\hline
		HTSK-LLM-KD & 78.63/78.52  & 75.21/75.12  & 3.42/3.40 \bigstrut[t]\\
		\hline
		HTSK-LLM-DKD & 79.10/79.08  & 75.60/75.62  & 3.50/3.46 \bigstrut[t]\\
		\hline
	\end{tabular}%
	\label{tab:addlabel}%
\end{table}

From the experiments results above we can conclude that knowledge distillation can improve the generalization ability of model. Next, we will explain how knowledge distillation can affect the performance of student model on the real-world dataset \textit{Cleveland heart disease}.
\subsection{Experimental Results and Analysis on \textit{Cleveland Heart Disease} Dataset}
We use a real-world dataset \textit{Cleveland heart disease} (\textit{CLE}) to demonstrate the performance of our model in detail. \textit{CLE} is about heart disease in Cleveland city of USA, which is built by University Hospital Zurich and Cleveland Clinic Foundation. \textit{CLE} is composed of 303 instances with 13 features, including chest pain type, resting blood pressure, resting electrocardiographic results, etc., and the output indicates what risk does the patient have with cardiac disease, which is an integer number ranging from 0 (not present) to 4: 0 for nil risk; 1 for low risk; 2 for potential risk; 3 for high risk; 4 for very high risk \cite{ref42}. In order to  clearly demonstrate the effect of knowledge distillation on the proposed model, we divide \textit{CLE} into three classes: 0 for zero-risk (nil risk), 1 for low-risk (low risk, potential risk and high risk) and 2 for high-risk (very high risk).
The experiments in Table~\ref{table3} shows that HTSK-LLM-DKD and HTSK-LLM-KD defeat other comparative methods and obtain the first best and the second best performance on \textit{CLE}, respectively. In addition, HTSK-LLM-DKD and HTSK-LLM-KD require fewer fuzzy rules (8.1 \& 8.6) and less running time (0.3168 \& 0.2782), which shows once again that the fuzzy dark knowledge greatly improves the generalization ability of model. In Table~\ref{table4}, HTSK-LLM-DKD achieves the best performance improvement with 3.50\% in $Accuracy$ and 3.46\% in $Weighted-F$, which are better than HTSK-LLM-KD with  3.42\% in $Accuracy$ and 3.40\% in $Weighted-F$, illustrating the great performance improvement brought by decoupled knowledge distillation once again, which is in accordance with the conclusions obtained in Section \uppercase\expandafter{\romannumeral4}-B.

\begin{figure}
	\centering
	\includegraphics[width=3.5in]{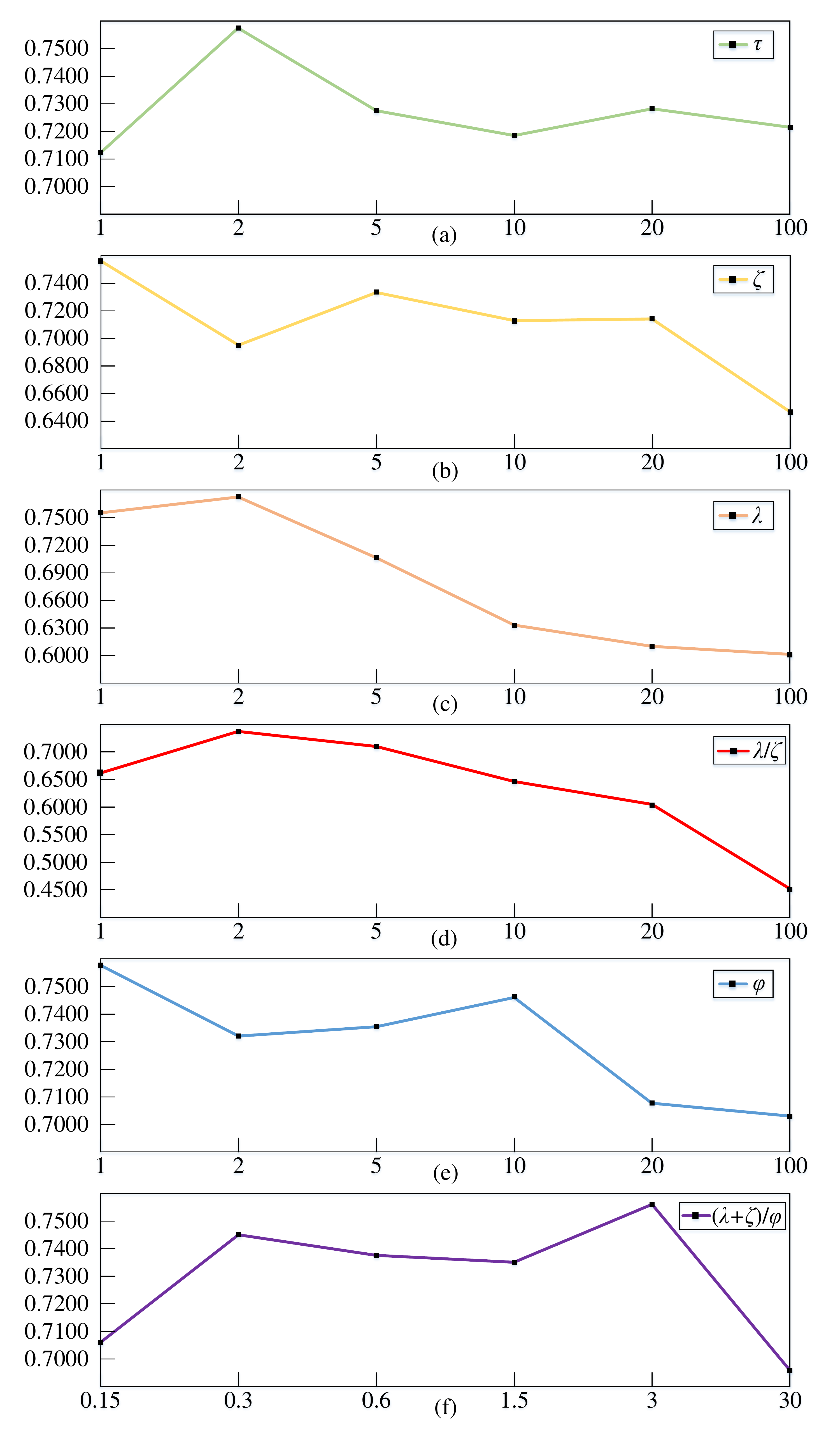}
	\caption{The effects of distillation parameters on \textit{Cleveland heart disease} dataset.}
	\label{fig_3}
\end{figure}
\subsection{Effectiveness of Decoupled Knowledge Distillation}

Fig.~\ref{fig_3} shows the effects of distillation parameters of the proposed HTSK-LLM-DKD on dataset \textit{CLE}. All distillation parameters are selected from $\lbrace1, 2, 5, 10, 20, 100\rbrace$, and it can be clearly seen that in the process of most parameter selection, the classification performance improves first and then degrades. The distillation temperature $\tau$ has an important effect on the softness of class label. Fig.~\ref{fig_3}(a) shows that distillation temperature $\tau$ in the range $\lbrace2, 20\rbrace$ may be a better choice. It means that the lower temperature can't effectively distill the similar information between classes, while the higher temperature destroys the prediction of various classes, so the temperature in the middle is the most appropriate. $\zeta$ and $\lambda$ show the proportion of target class knowledge and non-target class knowledge, respectively, as shown in Figs.~\ref{fig_3}(b) \& (c). We use $\lambda/\zeta$ to show the proportion of them. Fig.~\ref{fig_3}(d) reveals that $\lambda/\zeta$ within $\lbrace2, 5\rbrace$ may be a better option, which means that non-target class knowledge can better improve the efficiency of knowledge distillation, thus improving the performance of the model, but too much non-target class knowledge will also reduce the classification performance of the model, because the target class knowledge transfers the fuzzy dark knowledge about the sample difficulty. $\varphi$ in Fig.~\ref{fig_3}(e) exhibits the proportion of the knowledge contained in the ground-truth label, we use $(\lambda+\zeta)/\varphi$ to show the proportion of fuzzy dark knowledge and knowledge contained in the ground-truth label. Fig.~\ref{fig_3}(f) displays that $(\lambda+\zeta)/\varphi$ within $\lbrace0.3, 3\rbrace$ may be a better choice, which means that a appropriate amount of fuzzy dark knowledge can improve the classification performance. We can conclude that when transferring too little fuzzy dark knowledge, student model cannot be effectively guided by teacher model, and when transferring too much fuzzy dark knowledge, student model will be misled by the mistakes made by teacher model.

\begin{table}[htbp]
	\caption{Fuzzy Rule of HTSK-LLM-DKD on \textit{CLE} Dataset}
	\label{table5}
	\centering
	\footnotesize
	\tabcolsep=1.5pt
	\begin{tabular}{ll}
		\hline
		\multicolumn{2}{l}{Fuzzy Rule of HTSK-LLM-DKD} \bigstrut[t]\\
		\hline
		\multicolumn{2}{l}{\textbf{Rule 1:}} \bigstrut[t]\\
		\multicolumn{1}{l}{\multirow{4}[-12]{*}{\textbf{IF:}}} & the \emph{1st} feature is \emph{very high}, and \\
		& the \emph{2nd} feature is \emph{medium}, and \\
		& ......, and \\
		& the \emph{13th} feature is \emph{very high}. \\
		\multicolumn{1}{l}{\multirow{3}[-8]{*}{\textbf{THEN:}}} & the \emph{1st} output is 0.3429+0.3453$\emph{x}_1$+...+0.1834$\emph{x}_{13}$ = 0.1201, \\
		& the \emph{2nd} output is 0.6039+0.2608$\emph{x}_{1}$+...+0.5618$\emph{x}_{13}$ = 0.0863, \\
		& the \emph{3rd} output is -0.3570-0.0543$\emph{x}_{1}$+...-0.4343$\emph{x}_{13}$ = -0.1403. \\
		\multicolumn{1}{l}{\textbf{Rule 2:}} & \multicolumn{1}{l}{} \\
		\multicolumn{1}{l}{\multirow{4}[-12]{*}{\textbf{IF:}}} & the \emph{1st} feature is \emph{very high}, and \\
		& the \emph{2nd} feature is \emph{very low}, and \\
		& ......, and \\
		& the \emph{13th} feature is \emph{low}. \\
		\multicolumn{1}{l}{\multirow{3}[-8]{*}{\textbf{THEN:}}} & the \emph{1st} output is 0.4320+0.1222$\emph{x}_{1}$+...-0.1354$\emph{x}_{13}$ = 1.6256, \\
		& the \emph{2nd} output is 0.4737-0.1269$\emph{x}_{1}$+...+0.1912$\emph{x}_{13}$ = -5.2402, \\
		& the \emph{3rd} output is -0.5363+0.4600$\emph{x}_{1}$+...+0.0088$\emph{x}_{13}$ = 2.3668. \\
		\multicolumn{1}{l}{\textbf{Rule 3:}} & \multicolumn{1}{l}{} \\
		\multicolumn{1}{l}{\multirow{4}[-12]{*}{\textbf{IF:}}} & the \emph{1st} feature is \emph{very low}, and \\
		& the \emph{2nd} feature is \emph{very low}, and \\
		& ......, and \\
		& the \emph{13th} feature is \emph{medium}. \\
		\multicolumn{1}{l}{\multirow{3}[-8]{*}{\textbf{THEN:}}} & the \emph{1st} output is 0.8216+0.1555$\emph{x}_{1}$+...+0.6766$\emph{x}_{13}$ = -0.7301, \\
		& the \emph{2nd} output is 0.1020+0.1342$\emph{x}_{1}$+...+0.1751$\emph{x}_{13}$ = -0.1926, \\
		& the \emph{3rd} output is -0.5423-0.1844$\emph{x}_{1}$+...-1.1100$\emph{x}_{13}$ = 1.4580. \\
		\hline
	\end{tabular}%
	\label{tab:addlabel}%
\end{table}%
\subsection{Interpretability of HTSK-LLM-DKD on \textit{Cleveland Heart Disease} Dataset}

The benefit of fuzzy classifiers is that they may be articulated in terms of linguistic explanation. Here, we choose one trial with the best $Accuracy$ on \emph{CLE} among all experiment trials, and then, list in Table~\ref{table5} the corresponding antecedent parts and consequent parts of all fuzzy rules. Due to space limitation, we only present three features of \emph{CLE} dataset, \emph{i.e.}, the first, the second, and the last (\emph{13th}) features of all three fuzzy rules. For a random given datum $\mathbf{x}=(-0.2730,-0.6799,\ldots,-16.4620)^{T}$, we can observed from Table~\ref{table5} that, each fuzzy set A$_{i}^{k}$ for the rule $k$ and feature $i$ in the If-Part of HTSK-LLM-DKD can be interpreted with a possible linguistic phrase. In the Then-Part of HTSK-LLM-DKD, the absolute value of the outputs in Rules 2 \& 3 are much larger than that of Rule 1, which means that Rules 2 \& 3 play a more important role in HTSK-LLM-DKD. In Rules 2 \& 3, the \emph{3rd} output value is obviously the largest, after comprehensive calculation, we can conclude that the final prediction of HTSK-LLM-DKD is \emph{3rd} class, which means high-risk.
\section{Conclusion}
In this study, we mainly focus on how to integrate knowledge distillation and TSK fuzzy classifiers in a deeper level. Therefore, we propose a novel TSK fuzzy classifier based decoupling knowledge distillation denoted as HTSK-LLM-DKD, transferring fuzzy dark knowledge from High-order TSK fuzzy classifier to Low-order TSK fuzzy classifier. HTSK-LLM-DKD uses LLM-DKD to decouple the fuzzy dark knowledge from High-order TSK fuzzy classifier into target class knowledge and non-target class knowledge, then transfers to Low-order TSK fuzzy classifier more efficiently, obtaining better classification performance with high interpretability. Experimental results on the benchmarking datasets and a real dataset \textit{Cleveland heart disease} demonstrate its effectiveness.

HTSK-LLM-DKD has some improvements that deserve further study. First of all, we will conduct in-depth research on practical applications including epilepsy detection and movement prediction. Secondly, these state-of-the-art knowledge distillation methods that can be employed to distill fuzzy dark knowledge to TSK fuzzy classifier, which will also be the focus of our future research.

\end{document}